\documentclass[runningheads,dvipsnames]{llncs}

\usepackage{graphicx}
\usepackage[sort]{cite}
\usepackage{todonotes}
\usepackage{amsmath}
\usepackage{amssymb}
\usepackage{bbm}
\usepackage{adjustbox}
\usepackage{array,colortbl,multirow,multicol,booktabs,ctable}

\usepackage[title]{appendix}
\usepackage{setspace}
\usepackage{pbox}
\usepackage[most]{tcolorbox}
\usepackage{caption}

\definecolor{skillcolor}{rgb}{1.0, 0.95, 0.8}
\definecolor{descriptioncolor}{rgb}{0.85, 0.91, 0.99}
\definecolor{sentencecolor}{rgb}{0.84, 0.91, 0.83}
\definecolor{othersentencecolor}{gray}{.9}

\graphicspath{{images/}}

\usepackage{color}

\usepackage{threeparttable}
\usepackage[inline,shortlabels]{enumitem}
\usepackage{hyperref}
\usepackage{url}
\hypersetup{colorlinks=true,
    linkcolor=black,
    filecolor=black,      
    urlcolor=black,
    citecolor=black}

\makeatletter
\makeatletter
\def\Url@twoslashes{\mathchar`\/\@ifnextchar/{\kern-.2em}{}}
\g@addto@macro\UrlSpecials{\do\/{\Url@twoslashes}}
\makeatother

\usepackage[nameinlink]{cleveref}
\crefname{figure}{Fig.\hspace{-1pt}}{Figs.\hspace{-1pt}}
\Crefname{figure}{Figure}{Figures}
\crefname{equation}{Eq.\hspace{-1pt}}{Eqs.\hspace{-1pt}}
\Crefname{equation}{Equation}{Equations}
\crefname{section}{Section}{Sections}
\Crefname{section}{Section}{Sections}
\crefname{table}{Table}{Tables}
\crefname{appendix}{Appendix}{Appendices}
\Crefname{appendix}{Appendix}{Appendices}

\AtBeginEnvironment{subappendices}{\crefalias{section}{appendix}}

\newcommand{\eg}{e.g., }

\usepackage{color}

\begin{document}

\title{Extreme Multi-Label Skill Extraction Training using Large Language Models}
\author{Jens-Joris~Decorte\inst{1,2} \and
Severine~Verlinden\inst{2} \and
Jeroen~Van~Hautte\inst{2} \and
\\ Johannes~Deleu\inst{1}
\and
Chris~Develder\inst{1}
\and
Thomas~Demeester\inst{1}
}
\authorrunning{J.-J.~Decorte \emph{et al.}}

\institute{Ghent University -- imec, 9052 Gent, Belgium \\
\email{\{jensjoris.decorte, johannes.deleu, chris.develder, thomas.demeester\}@ugent.be}\\
\url{https://ugentt2k.github.io/}
\and
TechWolf, 9000 Gent, Belgium\\
\email{\{jensjoris, severine, jeroen\}@techwolf.ai} \\
\url{https://techwolf.ai/}}
\maketitle
\begin{abstract}

Online job ads serve as a valuable source of information for skill requirements, playing a crucial role in labor market analysis and e-recruitment processes.
Since such ads are typically formatted in free text, natural language processing (NLP) technologies are required to automatically process them.
We specifically focus on the task of detecting skills (mentioned literally, or implicitly described) and linking them to a large skill ontology, making it a challenging case of extreme multi-label classification (XMLC).
Given that there is no sizable labeled (training) dataset are available for this specific XMLC task, we propose techniques to leverage general Large Language Models (LLMs).
We describe a cost-effective approach to generate an accurate, fully synthetic labeled dataset for skill extraction, and present a contrastive learning strategy that proves effective in the task.
Our results across three skill extraction benchmarks show a consistent increase of between 15 to 25 percentage points in \textit{R-Precision@5}
compared to previously published results that relied solely on distant supervision through literal matches.

\end{abstract}

%-----------------------------------------
\section{Motivation and Related Work}
\label{sec:motivaton}
%=========================================

Job ads are published online on a daily basis. 
They contain valuable information about economic trends in the labor market, such as the evolution of skill demand in time.
Given that vacant jobs are advertized in unstructured text, we need automatic information extraction methods, \eg to extract such mentioned skills.
Such information extraction is crucial in labor market analysis and e-recruitment applications, including resume screening and job recommendation systems.
Thus it is unsurprising that in the last decade, the number of studies on skill extraction methods has increased tenfold~\cite{khaouja2021survey}.

\begin{figure}[t]
\centering
\includegraphics[width=0.95\textwidth]{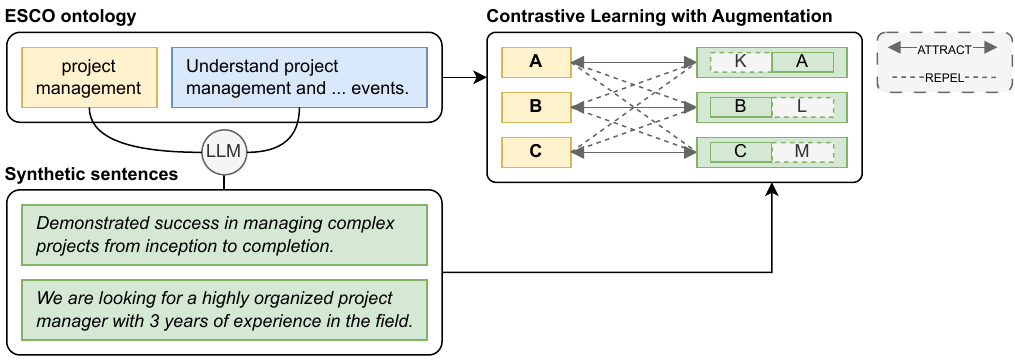}
\caption{\setlength\fboxsep{1pt}High-level illustration of our methodology. We start from the ESCO ontology, which contains \colorbox{skillcolor}{skills} and their \colorbox{descriptioncolor}{descriptions}. We use a LLM to generate \colorbox{sentencecolor}{synthetic sentences}. A bi-encoder is then optimized with contrastive learning to encode skill names and corresponding sentences in the same space. The sentences are augmented by randomly adding \colorbox{othersentencecolor}{other sentences} in front or behind them.} \label{fig:overview}
\end{figure}

Several works have simplified the skill \emph{extraction} problem to a pure \emph{detection} task, limited to identifying the text span expressing a skill.
Such a solution thus forgoes the normalization of synonyms and paraphrased skills toward an ontology of skill labels.
Yet, since the latter is crucial for a consistent and robust job market analysis,  skill extraction has been approached as an extreme multi-label classification (XMLC) task, requiring linking job ads to the relevant fine-grained skills out of a long list of skill labels in a given ontology~\cite{decorte2022design,bhola-etal-2020-retrieving,vermeerusing,zhang2023escoxlmr}.
These XMLC works suffer from the difficulty of constructing a qualitative training dataset: the huge number of labels (\eg around 14k in our case study) makes annotation for skill extraction slow, and annotators are not sufficiently knowledgeable in each domain of the ontology to perform the task accurately.

In this paper, we present the effective usage of Large Language Models (LLMs) to circumvent the difficulty of annotation by automatically generating synthetic training data for skill extraction.
The following paragraphs motivate our approach, which can be summarized as follows.
We use an LLM to generate training data for skill extraction, grounded in the ESCO ontology.
Based on this synthetic data, we optimize a model using contrastive learning to represent skill names and corresponding sentences close together in the same space.
Our key contribution is a novel end-to-end approach to training a skill extraction system, consisting of the cost-effective synthetic data generation and the contrastive learning procedure alongside an effective augmentation procedure.
The effectiveness of this method is compared against a distant supervision baseline on three skill extraction benchmarks.
We release a large dataset of 138K \textit{(skill, job ad sentence)} pairs, covering 99.5\% of the ESCO ontology.

\paragraph{Large Language Models for skill extraction:}

With the increasing capabilities of LLMs over the past years, we argue that the time and knowledge-intensive task of annotation for skill extraction has become more feasible.
The usage of LLMs for skill extraction has been proposed by SkillGPT~\cite{li2023skillgpt}, who use an LLM to summarize the skill requirements of job ads, after which the summary is embedded and compared with the ESCO skills through cosine similarity.
While effective, as the authors themselves point out, this method suffers from the fact that generating the summaries is non-deterministic (implying that results would not be fully reproducible).
Further, we note that needing to run each individual job ad at inference time through the LLM, also incurs non-negligible costs and latency.
From the latter observation, we argue that a more effective way is to use the LLM only at training time, and in particular focus on constructing a strong training dataset, that far exceeds the quality of training data that has been constructed manually before.
As illustrated in \cref{fig:overview}, we combine the domain knowledge captured in the ESCO ontology with the language understanding and generation capabilities of a large language model, to gather 138k pairs of skills and corresponding synthetic job ad sentences.
%%\footnote{The data will be released upon acceptance.}

\paragraph{Contrastive learning for skill extraction:}

The ESCO skill synonyms and descriptions have recently been used in a contrastive learning setup for skill extraction from German job ads~\cite{gnehm-etal-2022-fine}. 
The authors propose a two-stage approach for skill detection: first identifying skill mentions as text spans, followed by ranking the ESCO skills against those mentions.
The ranking is performed by a bi-encoder model, which is optimized through contrastive training on pairs of \textit{(skill, description)} and \textit{(skill, synonyms)} from ESCO~\cite{gnehm-etal-2022-fine}.
However, we believe this approach is sub-optimal for two reasons.
First, skills are often mentioned as longer implicit spans or even full sentences, which pose challenges to span detection approaches. 
Second, ranking ESCO skills solely based on the detected spans restricts the utilization of contextual information and the strong contextual representation capabilities of BERT-based models.
Based on these insights, we adopt a contrastive learning approach that operates directly on full sentences, by training a bi-encoder to represent sentences and their corresponding skill labels closely together in representation space.
This strategy draws on a recent contribution demonstrating the strength of applying contrastive learning on pairs of biomedical concept names, and their corresponding textual descriptions \cite{remy2022biolord}.

%----------------------------
\section{Methodology}
\label{sec:methodology}
%============================

We make use of the ESCO ontology, which is briefly described in \cref{section:ontology}.
\cref{section:synthetic} describes how the ESCO skills and their descriptions are presented to an LLM to generate a large-scale synthetic dataset of positive \textit{(skill, job ad sentence)} pairs.
Finally, this data is used in a contrastive learning procedure to optimize a bi-encoder for skill extraction, as detailed in \cref{section:contrastive}.

\vspace{-0.3cm}
\subsection{Skill ontology: ESCO}
\label{section:ontology}
%---------------------------------------------------------

For this work, we use version 1.1.0 of the \textit{European Skills, Competences, Qualifications and Occupations} (ESCO) ontology~\cite{ESCO}.
ESCO contains names and descriptions of over 3k occupations and almost 14k skills, in 28 different languages.
We focus on the English version of ESCO and we only rely on skills.
These skills have on average 7 additional English synonyms linked to them, but on first inspection, these are not always entirely accurate, and we decided not to use them for our experiments (although the proposed method can be applied directly to the skill names augmented with the synonyms). 
Each skill is furthermore accompanied by a textual description, on which our approach toward synthetic data generation strongly relies.

\vspace{-0.3cm}
\subsection{Synthetic Data Generation}
\label{section:synthetic}
%---------------------------------------------------------

Aiming for a contrastive learning strategy, we need a set of positive training pairs \textit{(skill, job ad sentence)}, for all skills in the considered label space (i.e., ESCO skills).
Rather than starting from sentences and prompting the LLM to generate skill labels (for which an iterative strategy such as \cite{deraedt2023idas} may be used), we start from a single ESCO skill and prompt the LLM to generate sentences \textit{as if} from job ads, that require that skill.
As such, we avoid the difficulty of aligning the LLM output with the ESCO skill label set.
The \emph{gpt-3.5-turbo-0301} model was used through the paid API provided by OpenAI. The model was chosen because of its intuitive prompting procedure and competitive cost.

\vspace{-0.2cm}
\paragraph{LLM prompt design:} Two prompts were compared (provided in \cref{app:prompts}). 
The first prompt requests a list of job ad sentences requiring a certain skill, with demonstrations for the skills \textit{Java} and \textit{project management}.
With this method, the LLM responded with a list of sentences for 73\% of the ESCO skills.
In the other cases, the response stated that not enough information about the skill was provided to generate accurate sentences, or that it would be highly unlikely that job ads request the skill.
The second prompt was obtained by extending the initial prompt to include the ESCO skill description, as well as a clarification that the requested job ads are \textit{hypothetical}.
This increased the number of responses with lists of sentences from 73\% to 99.5\%.
The accuracy of the generated data was manually assessed on a subset, and determined to be at around 88\% and 94\%, respectively, for the first and second prompts.
The final dataset was generated with the second prompt and consists of 10 synthetic sentences for 13,826 unique ESCO skills.\footnote{\url{https://huggingface.co/datasets/jensjorisdecorte/Synthetic-ESCO-skill-sentences}}
Some examples from this dataset are listed in \cref{app:examples}.

\vspace{-0.3cm}
\subsection{Contrastive Learning for Skill Extraction}
\label{section:contrastive}
%---------------------------------------------------------

For the contrastive learning, we use a bi-encoder architecture, in which pairs of skills and corresponding sentences are encoded by the same encoder-only transformer architecture~\cite{reimers2019sentence}.
We make use of \textit{multiple negatives ranking loss} with in-batch negatives, as proposed by \cite{Henderson2017EfficientNL}.
The setup is visualized in \cref{fig:overview}.
Inspired by the work in \cite{gnehm-etal-2022-fine}, we also train a model using \textit{(skill, ESCO description}) pairs, to compare the benefit of using the synthetic job ad sentences.
The trained bi-encoder is directly used for the task of skill extraction, by ranking all ESCO skills with respect to an input sentence, based on the cosine similarity of their representations.
In other words, no supervised fine-tuning is performed on the skill extraction task.

\paragraph{Augmentation:} We note that the synthetic sentences typically only discuss its linked skill, while real sentences can mention more skills.
We hypothesize that this setup limits the model's capability to reflect multiple skills in its embeddings, which would harm performance for skill extraction.
To this end, we introduce an augmentation strategy that randomly adds another sentence in front or behind each sentence during training.
The pairs now consist of a skill and two concatenated sentences, of which only one sentence relates to the skill.
We argue that this forces the model to represent both of the concatenated sentences to match with the correct skill in the batch.
This augmentation is visualized in \cref{fig:overview}.

%-----------------------------------------------
\section{Experimental Setup and Results}
\label{sec:experiments}
%===============================================

We start from a sentence-transformer model (\textit{all-mpnet-base-v2}\footnote{\url{https://huggingface.co/sentence-transformers/all-mpnet-base-v2}}) that was pretrained on over 1B English sentence pairs~\cite{reimers2019sentence}.
The synthetic training dataset is compared to directly using pairs of ESCO skills and their descriptions, with and without the proposed augmentation method.
For each dataset, we use the same hyperparameters, reported in \cref{app:training}.
The performance is evaluated against the \textsc{TECH} and \textsc{HOUSE} benchmarks provided in~\cite{decorte2022design}, and a proprietary test set \textsc{TECHWOLF}. Each benchmark consist of hundreds of job ad sentences annotated with corresponding ESCO skills.
Results are expressed in terms of mean reciprocal rank (MRR) and r-precision at 5 (RP@5), as in~\cite{decorte2022design}, shown in \cref{tab:results}.
RP@K is defined in \eqref{eq:RP}, where the quantity $Rel(n,k)$ is a binary indicator of whether the $k^\text{th}$ ranked label is a correct label for data sample $n$, and $R_n$ is the number of gold labels for sample $n$.

\vspace{-0.2cm}

\begin{equation}
    RP@K = \frac{1}{N} \sum\limits_{n=1}^{N} \frac{1}{\min(K, R_n)}\sum\limits_{k=1}^{K} Rel(n,k)
    \label{eq:RP}
\end{equation}

\vspace{-0.6cm}

\begin{table*}[bht]
\fontsize{9pt}{9pt}\selectfont
    \centering
    \begin{threeparttable}
    \begin{tabular}{l cc cc cc cc}
    \toprule
    & \multicolumn{2}{c}{\emph{TECH}} & \multicolumn{2}{c}{\emph{HOUSE}} & \multicolumn{2}{c}{\emph{TECHWOLF}} \\
    \cmidrule(lr){2-3} \cmidrule(lr){4-5} \cmidrule(lr){6-7}
    \textbf{Metric} & MRR & RP@5 & MRR & RP@5 & MRR & RP@5 \\
    \midrule
    Distant Supervision~\cite{decorte2022design} & 32.50 & 32.12 & 29.59 & 30.57 & 28.44 & 29.27\\
    \midrule
    \textit{all-mpnet-base-v2} & 38.76 & 39.60 & 26.27 & 26.17 & 29.58 & 33.48\\
    \midrule
    ESCO descriptions & 46.83 & 48.10 & 36.46 & 37.17 & 42.09 & 44.96 \\
    ESCO descriptions\tnote{\textsc{aug}} & \underline{48.46} & \underline{51.99} & 39.15 & \underline{42.35} & 44.13 & 45.87 \\
    GPT sentences & 48.33 & 48.80 & \underline{41.13} & 40.80 & \underline{46.50} & \underline{51.24} \\
    GPT sentences\tnote{\textsc{aug}}  & \textbf{52.85} & \textbf{54.62} & \textbf{42.75} & \textbf{45.74} & \textbf{52.55} & \textbf{54.57} \\
    \bottomrule
    \end{tabular}
    \begin{tablenotes}
    \small
    \item[\textsc{aug}]:  with the proposed augmentation.\\
    \end{tablenotes}
    
    \end{threeparttable}
    \caption{Performance of the different training regimes. The distant supervision method from \cite{decorte2022design} is reproduced. The \textit{all-mpnet-base-v2} model is directly evaluated without further training. Best results are \textbf{bold}, second best are \underline{underlined}.}
    \label{tab:results}
\end{table*}

\begin{figure}[ht]
\centering
\includegraphics[width=\textwidth]{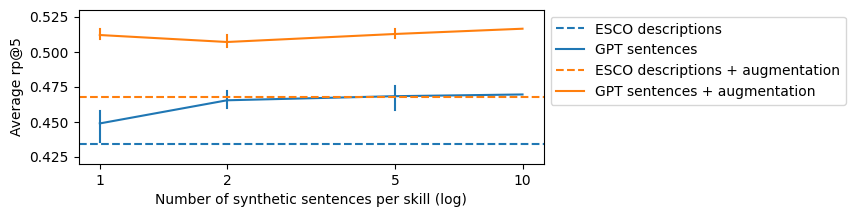}
\vspace{-0.5cm}
\caption{Average rp@5 across all benchmarks, for different amounts of sentences per skill during training. Note that just using one sentence per skill already outperforms the description-based training. The horizontal axis is in logarithmic scale. Error bars show the minimum and maximum performance of the runs.} \label{fig:abbl}
\end{figure}

Synthetic sentences yield superior performance to ESCO descriptions, and augmentation has a positive impact in all cases.
We analyze the performance as we randomly subsample the number of synthetic sentences per skill.
As shown in figure~\ref{fig:abbl}, more data helps, but even one synthetic sentence outperforms the description-based models.

\vspace{-0.5cm}

Note that the sentence representation of the model is simply an average of the embeddings of each token in the input.
This allows us to understand the relation between the predicted skills and the words in the input.
\cref{app:vis} contains a detailed explanation of how to do this, with some visual examples that help interpret the relation between skills and sentences in the model.

%----------------------------
\section{Conclusion}
\label{sec:conclusion}
%============================

This paper presents a cost-effective method for generating a comprehensive synthetic dataset of sentences, grounded in the ESCO ontology.
The size of this dataset surpasses any previously annotated dataset for skill extraction and covers 99.5\% of skills in ESCO.
We demonstrate that a bi-encoder can be optimized using a contrastive training procedure to effectively represent both skill names and corresponding sentences in close proximity within the same space.
This approach outperforms our distance supervision baseline by a large margin.
Additionally, we propose a simple augmentation method that enhances the resulting model quality.
We release the full dataset to foster future research in this area.

\section*{Ethical considerations}

It is crucial to carefully monitor and reduce any potential biases that might emerge from the data generation procedure.
Bias evaluation and mitigation strategies need to be in place to make sure that the final model does not reinforce unjust or discriminating outcomes.
Finally, any development of a skill extraction method should keep in mind the final application for which the extracted skills will serve.
Only with respect to this application can fairness be defined and evaluated.

\section*{Acknowledgments}

We thank the anonymous reviewers for their valuable feedback. This project was funded by the Flemish Government, through Flanders Innovation \& Entrepreneurship (VLAIO, project HBC.2020.2893).

%%\newpage
\bibliographystyle{splncs04}
\bibliography{mybib}

\begin{subappendices}
\renewcommand{\thesection}{\Alph{section}}

%%\newpage

\section{Examples of Synthetic Data}\label{app:examples}

We include the synthetic job ad sentences for two skills from ESCO. The following sentences were generated for the skill \texttt{interpret medical results} with description \textit{``Interpret, integrate and apply results of diagnostic imaging, laboratory tests and other investigations as part of the assessment of the client, in consultation with other healthcare practitioners.''}

\texttt{\small
\begin{enumerate}
    \item Applicants must be able to accurately interpret laboratory tests to assist in clinical decision making.
    \item We are currently seeking a professional who can accurately interpret medical results.
    \item We are looking for an experienced healthcare professional who can effectively communicate with others to provide quality care based on medical results.
    \item we are looking for a medical professional who can apply their expertise in interpreting complex medical results to improve patient outcomes.
    \item The position requires an understanding of how to integrate and apply diagnostic results in clinical practice.
    \item working experience in a clinical setting and interpreting medical test results is an advantage.
    \item we are seeking a candidate who can accurately interpret medical results and provide appropriate treatment recommendations.
    \item Must have experience interpreting medical results in a clinical setting.
    \item the candidate we are looking for must have the ability to integrate results of diagnostic tests and other investigations.
    \item excellent analytical and problem-solving skills are needed to be able to interpret medical results.
\end{enumerate}
}

\noindent Similarly, for the skill \texttt{shunt inbound loads} with description \textit{``Shunt inbound freight loads to and from railcars for inbound and outbound trains.''}, the synthetic sentences are:

\texttt{\small
\begin{enumerate}
    \item we are seeking a skilled laborer to shunt inbound loads efficiently and safely
    \item Qualified candidates should possess a strong understanding of shunting techniques as well as experience operating shunting equipment.
    \item we need someone who can shunt inbound loads quickly and accurately
    \item must have experience in shunting inbound freight loads
    \item Successful applicants will have a track record of accuracy and attention to detail while shunting inbound loads.
    \item We need someone who can effectively and efficiently shunt inbound loads to minimize downtime.
    \item the ability to shunt inbound loads is a must-have skill for this position
    \item shunting inbound loads is a physically demanding job
    \item We are looking for experienced operators with the ability to shunt inbound loads.
    \item Shunt drivers must be able to lift heavy loads and follow safety protocols.
\end{enumerate}
}

\section{Prompts}
\label{app:prompts}

We make use of the conversational ``user'' and ``assistant'' roles in the \emph{gpt-3.5-turbo-0301} model to integrate demonstrations into the prompt.
The information about the ESCO skill at hand needs to be pasted into the prompt, indicated by \texttt{\underline{\textit{variables}}} in the prompt.
The first version of our prompt is shown below:

\begin{tcolorbox}[fontupper=\footnotesize, colback=lightgray!20, boxrule=0.5pt, arc=4pt, boxsep=0pt, left=4pt, right=4pt, top=2pt, bottom=2pt, colframe=black, sharp corners]
\texttt{\textbf{User:}
Write two sentences from job ads that require the skill Java.\\
\\
\textbf{Assistant:} - experience with Java development, preferably web-based\\
- looking for a Java programmer this summer\\
\\
\textbf{User:}
Write two sentences from job ads that require the skill project management.\\
\\
\textbf{Assistant:}
- successful project managers are able to manage multiple tasks and deadlines simultaneously\\
- being able to effectively manage projects can give you valuable experience and skills\\
\\
\textbf{User:}
Write 10 sentences from job ads that require the skill \underline{\textit{skill}}}.
\end{tcolorbox}

\noindent The second version of the prompt is more lengthy.
It makes use of the ``system'' message to specify that the requested job ads are hypothetical.
The demonstrations follow a more structured format, and contain the ESCO description.
The response format of the ``assistant'' remains the same.

\begin{tcolorbox}[fontupper=\footnotesize, colback=lightgray!20, boxrule=0.5pt, arc=4pt, boxsep=0pt, left=4pt, right=4pt, top=2pt, bottom=2pt, colframe=black, sharp corners]
\texttt{\textbf{System:}
Respond with sentences from hypothetical job ads that require a certain skill, as asked by the user.\\
\\
\textbf{User:}
Number of sentences: 2\\
Skill: Java\\
Definition: The techniques and principles \{...\} in Java.\\
\\
\textbf{Assistant:} - experience with Java development, preferably web-based\\
- looking for a Java programmer this summer\\
\\
\textbf{User:}
Number of sentences: 2\\
Skill: project management\\
Definition: Understanding project management and \{...\} events.\\
\\
\textbf{Assistant:}
- successful project managers are able to manage multiple tasks and deadlines simultaneously\\
- being able to effectively manage projects can give you valuable experience and skills\\
\\
\textbf{User:}
Number of sentences: 10\\
Skill: \underline{\textit{skill}}\\
Definition: \underline{\textit{skill description}}}
\end{tcolorbox}

\newpage

\section{Training details}\label{app:training}

The contrastive training is implemented using the popular SBERT implementation~\cite{reimers2019sentence}.
We keep the default value of 20 for the ``scale'' hyperparameter \textit{alpha}.
We always train for 1 epoch.
The positive pairs are randomly shuffled into batches of 64.
We use the AdamW optimizer with a learning rate of 2e-5 and a ``WarmupLinear'' learning rate schedule with a warmup period of 5\% of the training data.
Automatic mixed precision was used to speed up training.
All experiments where performed using an Nvidia T4 GPU.

\section{Visualization of Predictions}\label{app:vis}

Table~\ref{tab:label-visualization} contains the top 3 predictions for two sentences that both contain multiple skill concepts.
The sentence representation of the model is simply an average of the embeddings of each token in the input.
This allows us to understand the relation between the predicted skills and the words in the input.
Concretely, we visualize the cosine similarity between the skill embedding and the embeddings of each of the tokens in the sentence.
This provides a visual interpretation of the relative importance of the input words with respect to the skills, which we find to be an interesting approach to understanding what the model has learned.

\begin{table}[htbp]
\fontsize{8pt}{8pt}
    \centering
\begin{tabular}{l|c}
\hline
\textbf{Label} & \textbf{Visualization} \\
\hline
\rowcolor{lightgray}\multicolumn{2}{c}{\textit{Coach IT teams with strong C++ skills}} \\
\hline
\textsc{C++} & \colorbox{red!6}{coach}\colorbox{red!24}{it}\colorbox{red!7}{teams}\colorbox{red!17}{with}\colorbox{red!33}{strong}\colorbox{red!71}{c}\colorbox{red!67}{\scriptsize+}\colorbox{red!66}{\scriptsize+}\colorbox{red!31}{skills} \\
\hline
\textsc{Microsoft Visual C++} & \colorbox{red!4}{coach}\colorbox{red!21}{it}\colorbox{red!7}{teams}\colorbox{red!13}{with}\colorbox{red!26}{strong}\colorbox{red!47}{c}\colorbox{red!47}{\scriptsize+}\colorbox{red!42}{\scriptsize+}\colorbox{red!22}{skills} \\
\hline
\textsc{coach employees} & \colorbox{red!53}{coach}\colorbox{red!5}{it}\colorbox{red!27}{teams}\colorbox{red!23}{with}\colorbox{red!19}{strong}\colorbox{red!5}{c}\colorbox{red!3}{\scriptsize+}\colorbox{red!4}{\scriptsize+}\colorbox{red!23}{skills} \\
\hline
\rowcolor{lightgray}\multicolumn{2}{c}{\textit{Onboard new colleagues and manage salaries}} \\
\hline
\textsc{introduce new employees} & \colorbox{red!40}{onboard}\colorbox{red!52}{new}\colorbox{red!35}{colleagues}\colorbox{red!23}{and}\colorbox{red!20}{manage}\colorbox{red!14}{salaries}\\
\hline
\textsc{determine salaries} & \colorbox{red!11}{onboard}\colorbox{red!16}{new}\colorbox{red!16}{colleagues}\colorbox{red!34}{and}\colorbox{red!28}{manage}\colorbox{red!54}{salaries}\\
\hline
\textsc{hire new personnel} & \colorbox{red!29}{onboard}\colorbox{red!41}{new}\colorbox{red!28}{colleagues}\colorbox{red!21}{and}\colorbox{red!15}{manage}\colorbox{red!15}{salaries}\\
\hline
\end{tabular}
\caption{Top 3 predicted skills for each sentence using the \textsc{GPT sentences\textsuperscript{aug}} model. For each skill, the sentence is visualized with the cosine similarity between the skill embedding and each token embedding. Higher cosine similarity is indicated by darker background color.}
\label{tab:label-visualization}
\end{table}

\end{subappendices}

\end{document}